\documentclass[10pt,twocolumn,letterpaper]{article}

\usepackage{cvpr}
\usepackage{times}
\usepackage{epsfig}
\usepackage{graphicx}
\usepackage{amsmath}
\usepackage{amssymb}
\usepackage{bm}
\usepackage{makecell}
\usepackage{multirow}
\usepackage{comment}
\usepackage{color}
\usepackage{rotating}
\usepackage{overpic}
\usepackage{xcolor}
\usepackage{colortbl}
\usepackage{hhline}

\usepackage[
   pagebackref=false,
   breaklinks=true,
   letterpaper=true,
   colorlinks,
   bookmarks=false]{hyperref}

\cvprfinalcopy
\def\cvprPaperID{3220}

\newif\ifarxiv
\arxivtrue

\def\Hline{\Xhline{2\arrayrulewidth}}     

\definecolor{mediumgreen}{RGB}{13,204,128}

\begin{document}
\title{Regularizing Deep Networks by Modeling and Predicting Label Structure}
\author{%
Mohammadreza Mostajabi \and Michael Maire \and Gregory Shakhnarovich\\
\vspace{-0.028\linewidth}\and%
Toyota Technological Institute at Chicago\\
{\tt\small\{mostajabi,mmaire,greg\}@ttic.edu}%
\vspace{-0.01\linewidth}%
}

\maketitle

\begin{abstract}
We construct custom regularization functions for use in supervised training
of deep neural networks.  Our technique is applicable when the ground-truth
labels themselves exhibit internal structure; we derive a regularizer by
learning an autoencoder over the set of annotations.  Training thereby becomes
a two-phase procedure.  The first phase models labels with an autoencoder.
The second phase trains the actual network of interest by attaching an
auxiliary branch that must predict output via a hidden layer of the
autoencoder.  After training, we discard this auxiliary branch.

We experiment in the context of semantic segmentation, demonstrating this
regularization strategy leads to consistent accuracy boosts over baselines,
both when training from scratch, or in combination with ImageNet pretraining.
Gains are also consistent over different choices of convolutional network
architecture.  As our regularizer is discarded after training, our method has
zero cost at test time; the performance improvements are essentially free.
We are simply able to learn better network weights by building an abstract
model of the label space, and then training the network to understand this
abstraction alongside the original task.

\end{abstract}

\section{Introduction}
\label{sec:introduction}

\begin{figure*}
   \begin{center}
      \includegraphics[width=0.925\linewidth]{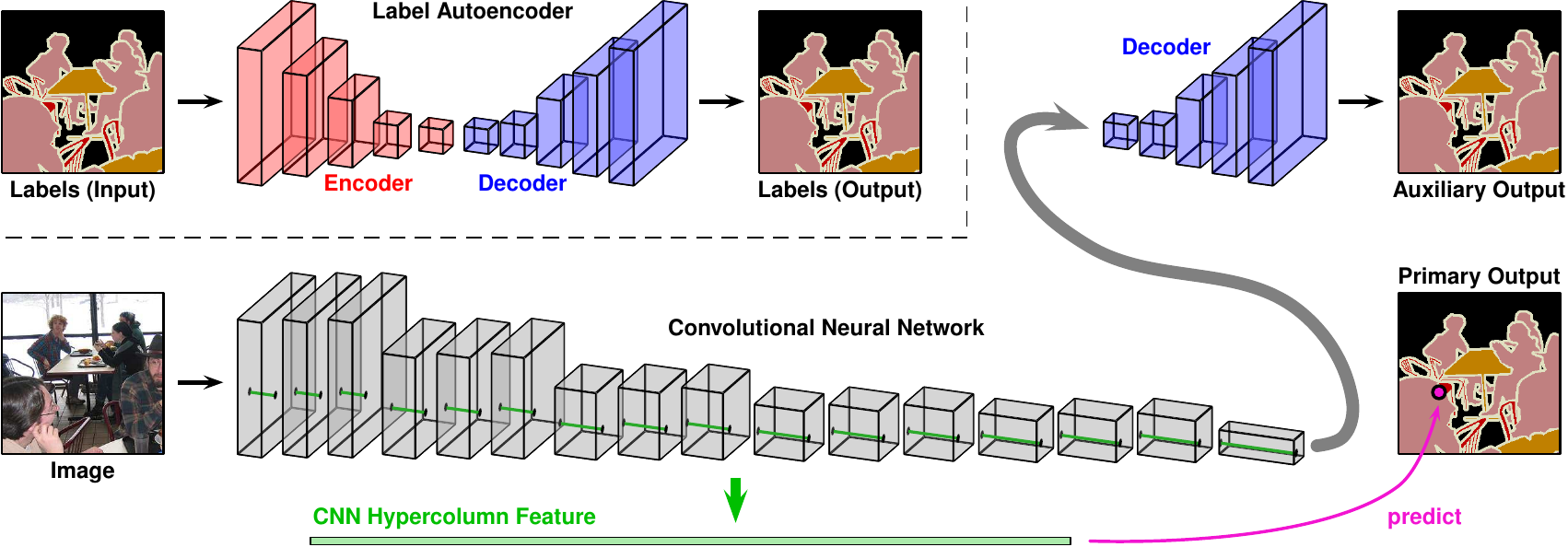}
   \end{center}
   \vspace{-0.01\linewidth}
   \caption{
      \textbf{Exploiting label structure when training semantic segmentation.}
      \emph{\textbf{Top:}}
      An initial phase looks only at the ground-truth annotation of training
      examples, ignoring the actual images.  We learn an autoencoder that
      approximates an identity function over segmentation label maps.  It is
      constrained to compress and reconstitute labels by passing them through
      a bottleneck connecting an encoder (red) and decoder (blue).
      \emph{\textbf{Bottom:}}
      The second phase trains a standard convolutional neural network (CNN)
      for semantic segmentation using hypercolumn~\cite{Hypercolumns,ZoomOut}
      features for per-pixel output.  However, we attach an auxiliary branch
      (and loss) that also predicts segmentation by passing through the decoder
      learned in the first phase.  After training, we discard this decoder
      branch, making the architecture appear standard.
   }
   \label{fig:overview}
   \vspace{-0.015\linewidth}
\end{figure*}

The recent successes of supervised deep learning rely on the availability of
large-scale datasets with associated annotations for training.  In computer
vision, annotation is a sufficiently precious resource that it is commonplace
to pretrain systems on millions of labeled ImageNet~\cite{ImageNet} examples.
These systems absorb a useful generic visual representation ability during
pretraining, before being fine-tuned to perform more specific tasks using
fewer labeled examples.

Current state-of-the-art semantic segmentation methods~\cite{DeepParsing,
DeepLab,PSPNet} follow such a strategy.  Its necessity is driven by the high
relative cost of annotating ground-truth for spatially detailed
segmentations~\cite{PASCAL,COCO}, and the accuracy gains achievable by
combining different data sources and label modalities during training.  A
collection of many images, coarsely annotated with a single label per image
(\eg ImageNet~\cite{ImageNet}), is still quite informative in comparison to a
smaller collection with detailed per-pixel label maps for each image (\eg
PASCAL~\cite{PASCAL} or COCO~\cite{COCO}).

We show that detailed ground-truth annotation of this latter form contains
additional information that existing schemes for training deep convolutional
neural networks (CNNs) fail to exploit.  By designing a new training procedure,
we are able to capture some of this information, and as a result increase
accuracy at test time.

Our method is orthogonal to recent efforts, discussed in
Section~\ref{sec:related}, on learning from images in an unsupervised or
self-supervised manner~\cite{
CtxEncoder,
Jigsaw,
ZIE:ECCV:2016,
LMS:ECCV:2016,
LMS:CVPR:2017,
ZIE:CVPR:2017,
DKD:ICLR:2017}.
It is not dependent upon the ability to utilize an external pool of data.
Rather, our focus on more efficiently utilizing provided labels makes our
contribution complementary to these other learning techniques.  Experiments
show gains both when training from scratch, and in combination with pretraining
on an external dataset.

Our innovation takes the form of a regularization function that is itself
learned from the training set labels.  This yields two distinct training
phases.  The first phase models the structure of the labels themselves by
learning an autoencoder.  The second phase follows the standard network
training regime, but includes an auxiliary task of predicting the output via
the decoder learned in the first phase.  We view this auxiliary branch as a
regularizer; it is only present during training.  Figure~\ref{fig:overview}
illustrates this scheme.

Section~\ref{sec:method} further details our approach and the intuition behind
it.  Our regularizer can be viewed as a requirement that the system understand
context, or equivalently, as a method for synthesizing context-derived labels
at coarser spatial resolution.  The auxiliary branch must predict this more
abstract, context-sensitive representation in order to successfully interface
with the decoder.

Experiments, covered in Section~\ref{sec:experiments}, focus on the PASCAL
semantic segmentation task.  We take baseline CNN architectures, the
established VGG~\cite{VGG} network and the state-of-the-art DenseNet~\cite{
DenseNet}, and report performance gains of enhancing them with our custom
regularizer during training.  Section~\ref{sec:experiments} also provides
ablation studies, explores an alternative regularizer implementation, and
visualizes representations learned by the label autoencoder.

Results demonstrate performance gains under all settings in which we applied
our regularization scheme: VGG or DenseNet, with or without data augmentation,
and with or without ImageNet pretraining.  Performance of a very deep DenseNet,
with data augmentation and ImageNet pretraining, is still further improved with
use of our regularizer during training.  Together, these results indicate that
we have discovered a new and generally applicable method for regularizing
supervised training of deep networks.  Moreover, our method has no cost at
test time; it produces networks architecturally identical to baseline designs.

Section~\ref{sec:conclusion} discusses implications of our demonstration that
it is possible to squeeze more benefit from detailed label maps when training
deep networks.  Our results open up a new area of inquiry on how best to build
datasets and design training procedures to efficiently utilize annotation.

\section{Related Work}
\label{sec:related}

The abundance of data, but more limited availability of ground-truth
supervision, has sparked a flurry of recent interest in developing
self-supervised methods for training deep neural networks.  Here, the idea is
to utilize a large reserve of unlabeled data in order to prime a deep network
to encode generally useful visual representations.  Subsequently, that network
can be fine-tuned on a novel target task, using actual ground-truth
supervision on a smaller dataset.  Pretraining on ImageNet~\cite{ImageNet}
currently yields such portable representations~\cite{DeCAF}, but lacks the
ability to scale without requiring additional human annotation effort.

Recent research explores a diverse array of data sources and tasks for
self-supervised learning.  In the domain of images, proposed tasks include
inpainting using context~\cite{CtxEncoder}, solving jigsaw puzzles~\cite{
Jigsaw}, colorization~\cite{ZIE:ECCV:2016,LMS:ECCV:2016,LMS:CVPR:2017},
cross-channel prediction~\cite{ZIE:CVPR:2017}, and learning a bidirectional
variant~\cite{DKD:ICLR:2017} of generative adversarial networks (GANs)~\cite{
GAN}.  In the video domain, recent works harness temporal coherence~\cite{
MCW:ICML:2009,JG:CVPR:2016}, co-occurrence~\cite{IZKA:ICLRworkshop:2016}, and
ordering~\cite{MZH:ECCV:2016}, as well as tracking~\cite{WG:ICCV:2015},
sequence modeling~\cite{SMS:ICML:2015}, and motion grouping~\cite{
PGDDH:CVPR:2017}.  Owens~\etal~\cite{OWMFT:ECCV:2016} explore cross-modality
self-supervision, connecting vision and sound.  Agrawal~\etal~\cite{
ANAML:NIPS:2016} and Nair~\etal~\cite{NCAIAML:ICRA:2017} examine settings in
which a robot learns to predict the visual effects of its own actions.

Training a network to perform ImageNet classification or a self-supervised
task, in addition to the task of interest, can be viewed as a kind of implicit
regularization constraint.  Zhang~\etal~\cite{ZLL:ICML:2016} explore explicit
auxiliary reconstruction tasks to regularize training.  However, they focus on
encoding and decoding image feature representations.  Our approach differs
entirely in the source of regularization.

Specifically, by autoencoding the structure of the target task labels, we
utilize a different reserve of information than all of the above methods.  We
design a new task, but whereas self-supervision formulates the new task on
external data, we derive the new task from the annotation.  This separation of
focus allows for possible synergistic combination of our method with
pretraining of either the self-supervised or supervised (ImageNet) variety.
Section~\ref{sec:experiments} tests the latter.

Another important distinction from recent self-supervised work is that, as
detailed in Section~\ref{sec:method}, we use a generic mechanism, based on
an autoencoder, for deriving our auxiliary task.  In contrast, the vast
majority of effort in self-supervision has relied on using domain-specific
knowledge to formulate appropriate tasks.  Inpainting~\cite{CtxEncoder},
jigsaw puzzles~\cite{Jigsaw}, and colorization~\cite{ZIE:ECCV:2016,
LMS:ECCV:2016,LMS:CVPR:2017} exemplify this mindset;
BiGANs~\cite{DKD:ICLR:2017} are perhaps an exception, but to date their results
compare less favorably~\cite{LMS:CVPR:2017}.

The work of Xie~\etal~\cite{XHT:ECCV:2016} shares similarities to our approach
along the aspect of modeling label space.  However, they focus on learning a
shallow corrective model that essentially denoises a predicted label map using
center-surround filtering.  In contrast, we build a deep model of label space.
Also, unlike~\cite{XHT:ECCV:2016}, our approach has no test-time cost, as we
impose it only as a regularizer during training, rather than as an ever-present
denoising layer.

Inspiration for our method traces back to the era of vision prior to the
pervasive use of deep learning.  It was once common to consider context as
important~\cite{Torralba:IJCV:2003}, reason about object parts, co-occurrence,
and interactions~\cite{DRF:IJCV:2011}, and design graphical models to capture
such relationships~\cite{STFW:ICCV:2005}.  We refer to only a few sample
papers as fully accounting for a decade of computer vision research is not
possible here.  In the following section, we open a pathway to pull such
thinking about compositional scene priors into the modern era: simply
learn, and employ, a deep model of label space.

\section{Method}
\label{sec:method}

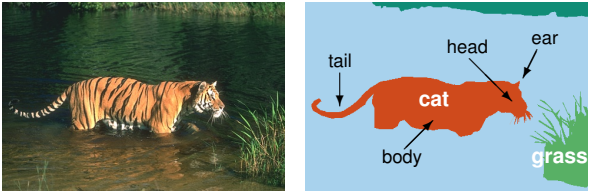
\begin{figure}[t]
   \setlength\fboxsep{0pt}
   \begin{center}
      \begin{minipage}[t]{0.01\linewidth}
         \vspace{0pt}
      \end{minipage}
      \hfill
      \begin{minipage}[t]{0.45\linewidth}
         \vspace{-1pt}
         \begin{center}
            \includegraphics[width=1.0\linewidth]{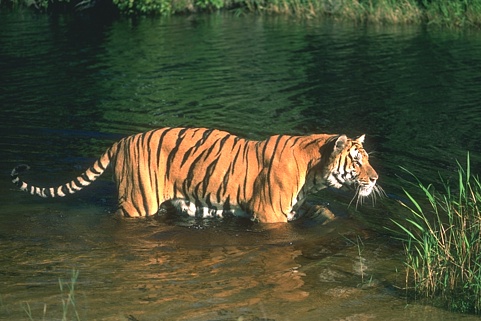}
         \end{center}
      \end{minipage}
      \hfill
      \begin{minipage}[t]{0.45\linewidth}
         \vspace{0pt}
         \begin{center}
            \begin{overpic}[width=1.0\linewidth,grid=false]{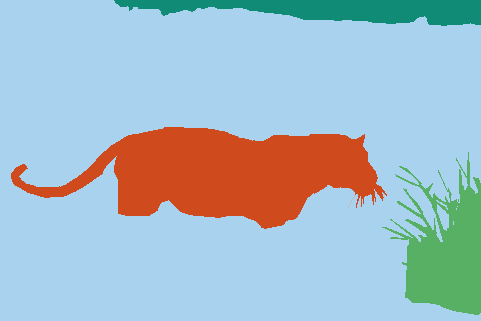}
               \put(40,30){\textcolor{white}{\footnotesize{\textsf{\textbf{cat}}}}}
               \put(80,10){\textcolor{white}{\footnotesize{\textsf{\textbf{grass}}}}}
               \put(8,44){\textcolor{black}{\scriptsize{\textsf{tail}}}}
               \put(12,42){\color{black}\vector(0,-4){12}}
               \put(80,52){\textcolor{black}{\scriptsize{\textsf{ear}}}}
               \put(82,50){\color{black}\vector(-1,-2){5}}
               \put(50,49){\textcolor{black}{\scriptsize{\textsf{head}}}}
               \put(58,47){\color{black}\vector(1,-1){15}}
               \put(27,10){\textcolor{black}{\scriptsize{\textsf{body}}}}
               \put(35,15){\color{black}\vector(1,1){10}}
            \end{overpic}
         \end{center}
      \end{minipage}
      \hfill
      \begin{minipage}[t]{0.01\linewidth}
         \vspace{0pt}
      \end{minipage}
   \end{center}
   \vspace{-0.01\linewidth}
   \caption{
      \textbf{Informative structure in annotation.}
      The shape of labeled semantic regions hints at unlabeled parts (black
      arrows).  Object co-occurrence provides a prior on scene composition.
   }
   \label{fig:tiger}
   \vspace{-0.03\linewidth}
\end{figure}

Figure~\ref{fig:tiger} is a useful aid in explaining the intuition behind the
regularization scheme outlined in Figure~\ref{fig:overview}.  Suppose we want
to train a CNN to recognize and segment cats, but our limited training set
consists only of tigers.  It is conceivable that the CNN will learn an
equivalence between black and orange striped texture and the cat category, as
such association suffices to classify every pixel on a tiger.  It thus
overfits to the tiger subclass and fails when tested on images of house cats.
This behavior could arise even if trained with detailed supervision of the
form shown in Figure~\ref{fig:tiger}.

Yet, the semantic segmentation ground-truth suggests to any human that texture
should not be the primary criteria.  There are no stripes in the annotation.
Over the entire training set, regions labeled as cat share a distinctive shape
that deforms in a manner suggestive of unlabeled parts (\eg head, body, tail,
ear).  The presence or absence of other objects in the scene may also provide
contextual cues as to the chance of finding a cat.  How can we force the CNN to
notice this wealth of information during training?

We could consider treating the ground-truth label map as an image, and
clustering local patches.  The patch containing the skinny tail would fall in a
different cluster than that containing the pointy ear.  Adding the cluster
identities as another semantic label, and requiring the CNN to predict them,
would force the CNN to differentiate between the tail and ear by developing a
representation of shape.  This clustering approach is reminiscent of
Poselets~\cite{BM:ICCV:2009,BMBM:ECCV:2010}.

Following this strategy, we would need to hand-craft another scheme for
capturing object co-occurrence relations, perhaps by clustering descriptors
spanning a larger spatial extent.  We would prefer a general means of capturing
features of the ground-truth annotations, and one not limited to a few
hand-selected characteristics.  Fortunately, deep networks are a suitable
general tool for building the kind of abstract feature hierarchy we desire.

\begin{figure}[t]
   \begin{center}
      \begin{scriptsize}
      \setlength\tabcolsep{4.5pt}
      \begin{tabular}{@{}c|c|c@{}}
      \footnotesize{Layers} & \footnotesize{DenseNet-67} & \footnotesize{DenseNet-121}\\
      \Hline
         & &\\[-7pt]
         Convolution
         &
            $\begin{bmatrix}
               3 \times 3\ \text{conv, stride 2}\\
               3 \times 3\ \text{conv}\\
               3 \times 3\ \text{conv}\\
            \end{bmatrix} \times 1$
         &
            $\begin{matrix}
               7 \times 7\ \text{conv,}\\
               \text{stride 2}\\
            \end{matrix}$
         \\[11pt]
      \hline
         Pooling &
            \multicolumn{2}{c}{$3 \times 3$ max pool, stride 2}\\
      \hline
         & &\\[-7pt]
         Dense Block (1)
         &
            $\begin{bmatrix}
               1 \times 1\ \text{conv}\\
               3 \times 3\ \text{conv}\\
            \end{bmatrix} \times 6$
         &
            $\begin{bmatrix}
               1 \times 1\ \text{conv}\\
               3 \times 3\ \text{conv}\\
            \end{bmatrix} \times 6$
         \\[5pt]
      \hline
         \multirow{2}{*}{Transition Layer (1)}
            & \multicolumn{2}{c}{$1 \times 1$ conv}\\
            \cline{2-3}
            & \multicolumn{2}{c}{$2 \times 2$ average pool, stride 2}\\
      \hline
         & &\\[-7pt]
         Dense Block (2)
         &
            $\begin{bmatrix}
               1 \times 1\ \text{conv} \\
               3 \times 3\ \text{conv}\\
            \end{bmatrix} \times 8$
         &
            $\begin{bmatrix}
               1 \times 1\ \text{conv} \\
               3 \times 3\ \text{conv}\\
            \end{bmatrix} \times 12$
         \\[5pt]
      \hline
         \\[-8pt]
         \multirow{2}{*}{Transition Layer (2)}
            & \multicolumn{2}{c}{$1 \times 1$ conv}\\
            \cline{2-3}
            & \multicolumn{2}{c}{$2 \times 2$ average pool, stride 2}\\
      \hline
         & &\\[-7pt]
         Dense Block (3)
         &
            $\begin{bmatrix}
               1 \times 1\ \text{conv} \\
               3 \times 3\ \text{conv}\\
            \end{bmatrix} \times 8$
         &
            $\begin{bmatrix}
               1 \times 1\ \text{conv} \\
               3 \times 3\ \text{conv}\\
            \end{bmatrix} \times 24$
         \\[5pt]
      \hline
         \multirow{2}{*}{Transition Layer (3)}
            & \multicolumn{2}{c}{$1 \times 1$ conv}\\
            \cline{2-3}
            & \multicolumn{2}{c}{$2 \times 2$ average pool, stride 2}\\
      \hline
         & &\\[-7pt]
         Dense Block (4)
         &
            $\begin{bmatrix}
               1 \times 1\ \text{conv}\\
               3 \times 3\ \text{conv}\\
            \end{bmatrix} \times 8$
         &
            $\begin{bmatrix}
               1 \times 1\ \text{conv}\\
               3 \times 3\ \text{conv}\\
            \end{bmatrix} \times 16$
         \\[5pt]
      \hline
      \end{tabular}
      \end{scriptsize}
      \vspace{0.01\linewidth}
      \caption{
         \textbf{DenseNet architectural specifications.}
      }
      \label{fig:densenet}
   \end{center}
   \vspace{-0.10\linewidth}
\end{figure}

\begin{figure*}
   \begin{center}
      \includegraphics[width=0.925\linewidth]{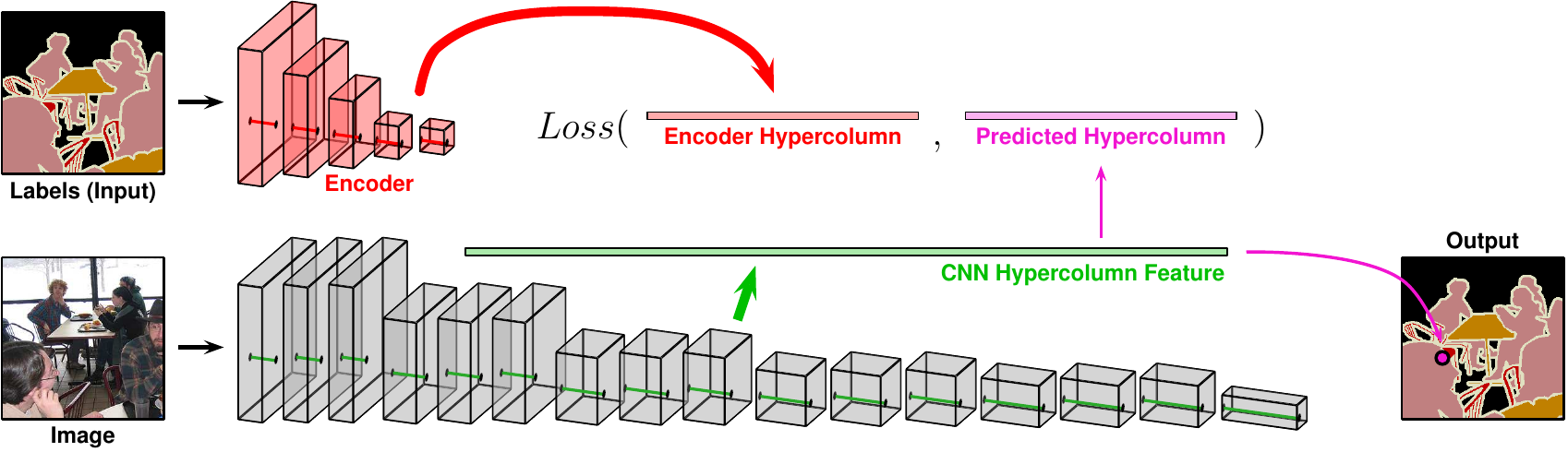}
   \end{center}
   \vspace{-0.01\linewidth}
   \caption{
      \textbf{Alternative regularization scheme.}
      Instead of predicting a representation to pass through the decoder, as
      in Figure~\ref{fig:overview}, we can train with an auxiliary regression
      problem.  We place a loss on directly predicting activations produced by
      the hidden layers of the encoder.
   }
   \label{fig:alternative}
   \vspace{-0.015\linewidth}
\end{figure*}

\subsection{Modeling Labels}
\label{sec:modeling}

Specifically, as shown in Figure~\ref{fig:overview}, we train an autoencoder
on the ground-truth label maps.  This autoencoder consumes a semantic
segmentation label map as input and attempts to replicate it as output.  By
virtue of being required to pass through a small bottleneck representation,
the job of the autoencoder is nontrivial.  It must compress the label map into
the bottleneck representation.  This compression constraint will (ideally)
force the autoencoder to discover and implicitly encode parts and contextual
relationships.

Ground-truth semantic segmentation label maps are simpler than real images, so
this autoencoder need not have as high of a capacity as a network operating on
natural images.  We use a relatively simply autoencoder architecture,
consisting of a mirrored encoder and decoder, with no skip connections.  The
encoder is a sequence of five $3 \times 3$ convolutional layers, with
$2 \times 2$ max-pooling between them.  The decoder uses $2\mathrm{x}$
upsampling followed by $3 \times 3$ convolution.  As a default, we set each
layer to have $32$-channels.  We also experiment with some higher-capacity
variants:
\begin{itemize}
   \setlength{\itemsep}{-2pt}
   \item{conv1: $32$-channels;~~conv2-5: $128$ channels each}
   \item{conv1: $32$;~~conv2-4: $128$;~~conv5: $256$ channels}
\end{itemize}
These channel progressions are for the encoder; the decoder uses the same in
reverse order.  We refer to these three autoencoder variants by the number
of channels in their respective bottleneck layers ($32$, $128$, or $256$).

\subsection{Baseline CNN Architectures}
\label{sec:architectures}

Convolutional neural networks for image classification gradually reduce
spatial resolution with depth through a series of pooling layers~\cite{
AlexNet,VGG,ResNet,DenseNet}.  As the semantic segmentation task requires
output at fine spatial resolution, some method of preserving or recovering
spatial resolution must be introduced into the architecture.  One option is
to gradually re-expand spatial resolution via upsampling~\cite{UNet,SegNet}.
Other approaches utilize some form of skip-connection to forward spatially
resolved features from lower layers of the network to the final layer~\cite{
FCN,Hypercolumns,ZoomOut}.  Dilated~\cite{YK:ICLR:2016} or atrous
convolutions~\cite{DeepLab} can also be mixed in.  Alternatively, the
basic CNN architecture can be reformulated in a multigrid setting~\cite{
KMY:CVPR:2017}.

Our goal is to examine the effects of a regularization scheme in isolation
from major architectural design changes.  Hence, we choose hypercolumn~\cite{
Hypercolumns,ZoomOut} CNN architectures as a primary basis for experimentation,
as they are are minimally separated from the established classification
networks in design space.  They also offer the added advantage of having
readily available ImageNet pretrained models, easing experimentation in this
setting.

We consider hypercolumn variants of VGG-16~\cite{VGG} and DenseNet~\cite{
DenseNet}.  These variants simply upsample and concatenate features from
intermediate network layers for use in predicting semantic segmentation.  As
shown in Figure~\ref{fig:overview}, this can equivalently be viewed as
associating with each spatial location a feature formed by concatenating a
local slice of every CNN layer.  The label of the corresponding pixel in the
output is predicted from that feature.

VGG-16 is widely used, while DenseNet~\cite{DenseNet} represents the latest
high-performance evolution of ResNet~\cite{ResNet}-like designs.  We use
67-layer and 121-layer DenseNets with the architectural details specified in
Figure~\ref{fig:densenet}.  The 67-layer net uses a channel growth rate of
$48$, while the 121-layer network, the same as in~\cite{DenseNet}, uses a
growth rate of $32$.  We work with $256 \times 256$ input and output spatial
resolutions in both CNNs and our label autoencoder.

\subsection{Regularization via Label Model}
\label{sec:regularization}

As shown by the large gray arrow in Figure~\ref{fig:overview}, we impose our
regularizer by connecting a CNN (\eg VGG or DenseNet) to the decoder portion of
our learned label autoencoder.  Importantly, the \emph{decoder parameters
are frozen during this training phase}.  The CNN now has two tasks, each with
an associated loss, to perform during training.  As usual, it must predict
semantic segmentation using hypercolumns.  It must also predict the same
semantic segmentation via an auxiliary path through the decoder.
Backpropagation from losses along both paths influences CNN parameter updates.
Though they participate in one of these paths, parameters internal to the
decoder are never updated.

We connect VGG-16 or DenseNet to the decoder by predicting input for the
decoder from the output of the penultimate CNN layer prior to global pooling.
This is the second-to-last convolutional layer, and is selected because its
spatial resolution matches that of the expected decoder input.  The prediction
itself is made via a new $1 \times 1$ convolutional layer, dedicated for that
purpose.

If the label autoencoder learns useful abstractions, requiring the CNN
to work through the decoder ensures that it learns to work with those
abstractions.  The hypercolumn pathway allows the CNN to make direct
predictions, while the decoder pathway ensures that the CNN has
``good reasons'' or a high-level abstract justification for its predictions.

Assuming autoencoder layers gradually build-up good abstractions, there
exist alternative methods of connecting it as a regularizer.
Figure~\ref{fig:alternative} diagrams one such alternative.  Here, we ask the
CNN to directly predict the feature representation built by the label encoder.
Encoder parameters are, of course, frozen here.  An auxiliary layer attempts to
predict the encoder hypercolumn from the CNN hypercolumn at the corresponding
spatial location.  The CNN must also still solve the original semantic
segmentation task.

As Section~\ref{sec:experiments} shows, this alternative scheme works well,
but not quite as well as using the decoder pathway.  Using the decoder is
also appealing for more reasons than performance alone.  Defining an auxiliary
loss in terms of decoder semantic segmentation output is more interpretable
than defining it in terms of mean square error (MSE) between two hypercolumn
features.  Moreover, the decoder output is visually interpretable; we can see
the semantic segmentation predicted by the CNN via the decoder.

\section{Experiments}
\label{sec:experiments}

The PASCAL dataset~\cite{PASCAL} serves as our experimental testbed.  We
follow standard procedure for semantic segmentation, using the official PASCAL
2012 training set, and reporting performance in terms of mean intersection over
union (mIoU) on the validation set (as validation ground-truth is publicly
available).  We explore both our decoder- and encoder-based regularization
schemes in combination with multiple choices of base network, data
augmentation, and pretraining.  When applying the encoder as a regularizer,
we task the CNN with predicting the concatenation of the encoder's activations
in its conv1 and conv3 layers.

\subsection{Setup}
\label{sec:setup}

All experiments are done in PyTorch~\cite{PyTorch}, using the Adam~\cite{Adam} update
rule when training networks.  Models trained from scratch use a batch size of
$12$ and learning rate of $\text{1}e^{-4}$ which after $80$ epochs decreased to
$\text{1}e^{-5}$ for an additional $20$ epochs.  For the case of ImageNet
pretrained models, we normalize hypercolumn features such that they have
zero-mean and unit-variance.  We keep the deep network weights frozen and
train the classifier for $10$ epochs with learning rate of $\text{1}e^{-4}$.
Then we decrease the learning rate to $\text{1}e^{-5}$ and train end-to-end for
additional $40$ epochs.

Data augmentation, when used, includes: a crop of random size in the
(0.08 to 1.0) of the original size and a random aspect ratio of 3/4 to 4/3 of
the original aspect ratio, which is finally resized to create a $256 \times 256$
image.  Plus random horizontal flip.  Pretrained models are based
on the PyTorch torchvision library~\cite{torchvision}.

We use cross-entropy loss on auxiliary regularization branches, except where
indicated by a superscript $\dagger$ in results tables.  For these experiments,
we use MSE loss.

\begin{table}[t]
   \definecolor{gray}{rgb}{0.33,0.33,0.33}
   \begin{center}
      \begin{small}
      \setlength\tabcolsep{3.5pt}
      \begin{tabular}{@{}l|c|c|>{\columncolor{white}[\tabcolsep][0pt]}r@{}}
      Architecture                        & Data-Aug?                & Auxiliary Regularizer                               & mIoU\\%
      \Hline%
                                          & \cellcolor{gray!15}{no}  & \cellcolor{gray!15}\textcolor{gray}{\textbf{none}}  & \cellcolor{gray!15}\textcolor{gray}{\textbf{37.3}}\\
      \multirow{2}{*}{VGG-16}             & no                       & Encoder (conv1 \& conv3)$^\dagger$                  & 41.1\\
      \multirow{2}{*}{%
         \footnotesize{~~-hypercolumn}}   & no                       & \textbf{Decoder (32 channel)}$^\dagger$             & \textbf{42.4}\\
                                          \hhline{~---}
                                          & \cellcolor{gray!15}{yes} & \cellcolor{gray!15}\textcolor{gray}{\textbf{none}}  & \cellcolor{gray!15}\textcolor{gray}{\textbf{55.2}}\\
                                          & yes                      & \textbf{Decoder (128 channel)}                      & \textbf{57.1}\\
      \Hline
      \multirow{2}{*}{%
         VGG-16\footnotesize{-FCN8s}}     & \cellcolor{gray!15}{yes} & \cellcolor{gray!15}\textcolor{gray}{\textbf{none}}  & \cellcolor{gray!15}\textcolor{gray}{\textbf{51.5}}\\
                                          & yes                      & \textbf{Decoder (128 channel)}                      & \textbf{54.1}\\
      \Hline
                                          & \cellcolor{gray!15}{no}  & \cellcolor{gray!15}\textcolor{gray}{\textbf{none}}  & \cellcolor{gray!15}\textcolor{gray}{\textbf{40.5}}\\
                                          & no                       & Encoder (conv1 \& conv3)$^\dagger$                  & 44.0\\
                                          & no                       & \textbf{Decoder (32 channel)}$^\dagger$             & \textbf{45.2}\\
      \multirow{1}{*}{DenseNet-67}        & no                       & Decoder (128 channel)                               & 42.5\\
                                          \hhline{~---}
      \multirow{1}{*}{%
         \footnotesize{~~-hypercolumn}}   & \cellcolor{gray!15}{yes} & \cellcolor{gray!15}\textcolor{gray}{\textbf{none}}  & \cellcolor{gray!15}\textcolor{gray}{\textbf{58.8}}\\
                                          & yes                      & Decoder (32 channel)                                & 59.4\\
                                          & yes                      & \textbf{Decoder (128 channel)}                      & \textbf{60.6}\\
                                          & yes                      & Decoder (256 channel)                               & 59.8\\
      \Hline
      \end{tabular}
      \end{small}
      \vspace{-0.01\linewidth}
      \caption{
         \textbf{PASCAL mIoU without ImageNet pretraining.}
         In each experimental setting (choice of architecture, and presence or
         absence of data augmentation), training with any of our regularizers
         improves performance over the baseline (shown in gray).
      }
      \label{tab:pascal_scratch}
   \end{center}
   \vspace{-0.065\linewidth}
\end{table}

\begin{table}[t]
   \begin{center}
      \begin{small}
      \setlength\tabcolsep{7.5pt}
      \begin{tabular}{@{}l|c|c|>{\columncolor{white}[\tabcolsep][0pt]}r@{}}
      Architecture                        & Data-Aug                 & Auxiliary Regularizer                               & mIoU\\
      \Hline
      &\cellcolor{gray!15}{}
      &\cellcolor{gray!15}{}
      &\cellcolor{gray!15}{}\\[-9pt]
                                          & \cellcolor{gray!15}{yes} & \cellcolor{gray!15}\textcolor{gray}{\textbf{none}}  & \cellcolor{gray!15}\textcolor{gray}{\textbf{58.8}}\\
      \multirow{1}{*}{DenseNet-67}        & yes                      & \textbf{Decoder (128 channel)}                      & \textbf{60.6}\\
      \multirow{1}{*}{%
         \footnotesize{~~-hypercolumn}}   & yes                      & Unfrozen Decoder                                    & 60.2\\
                                          & yes                      & Random Init.~Decoder                                & 58.8\\
      \Hline
      \end{tabular}
      \end{small}
      \vspace{-0.01\linewidth}
      \caption{
         \textbf{Ablation study.}
         PASCAL mIoU deteriorates if the decoder parameters are not held fixed
         while training the main CNN.
      }
      \label{tab:pascal_ablation}
   \end{center}
   \vspace{-0.065\linewidth}
\end{table}

\begin{table}[t]
   \begin{center}
      \begin{small}
      \setlength\tabcolsep{6pt}
      \begin{tabular}{@{}l|c|c|>{\columncolor{white}[\tabcolsep][0pt]}r@{}}
      Architecture                        & Data-Aug?                & Auxiliary Regularizer                               & mIoU\\
      \Hline
      &\cellcolor{gray!15}{}
      &\cellcolor{gray!15}{}
      &\cellcolor{gray!15}{}\\[-9pt]
      \multirow{1}{*}{VGG-16}             & \cellcolor{gray!15}{no}  & \cellcolor{gray!15}\textcolor{gray}{\textbf{none}}  & \cellcolor{gray!15}\textcolor{gray}{\textbf{67.1}}\\
      \multirow{1}{*}{%
         \footnotesize{~~-hypercolumn}}   & no                       & \textbf{Decoder (32 channel)}                       & \textbf{68.8}\\
      \Hline
      &\cellcolor{gray!15}{}
      &\cellcolor{gray!15}{}
      &\cellcolor{gray!15}{}\\[-9pt]
      \multirow{1}{*}{DenseNet-121}       & \cellcolor{gray!15}{yes} & \cellcolor{gray!15}\textcolor{gray}{\textbf{none}}  & \cellcolor{gray!15}\textcolor{gray}{\textbf{71.6}}\\
      \multirow{1}{*}{%
         \footnotesize{~~-hypercolumn}}   & yes                      & \textbf{Decoder (128 channel)}                      & \textbf{71.9}\\
      \Hline
      &\cellcolor{gray!15}{}
      &\cellcolor{gray!15}{}
      &\cellcolor{gray!15}{}\\[-9pt]
      \multirow{1}{*}{ResNet-101}         & \cellcolor{gray!15}{yes} & \cellcolor{gray!15}\textcolor{gray}{\textbf{none}}  & \cellcolor{gray!15}\textcolor{gray}{\textbf{75.4}}\\
      \multirow{1}{*}{%
         \footnotesize{~~-PSPNet}}        & yes                      & \textbf{Decoder (128 channel)}                      & \textbf{75.9}\\
      \Hline
      \end{tabular}
      \end{small}
      \vspace{-0.01\linewidth}
      \caption{
         \textbf{PASCAL mIoU with ImageNet pretraining.}
      }
      \label{tab:pascal_imagenet}
   \end{center}
   \vspace{-0.065\linewidth}
\end{table}

\begin{table}[t]
   \begin{center}
      \begin{small}
      \setlength\tabcolsep{7.5pt}
      \begin{tabular}{@{}l|c|c|>{\columncolor{white}[\tabcolsep][0pt]}r@{}}
      Architecture                        & Data-Aug                 & Auxiliary Regularizer                               & mIoU\\
      \Hline
      &\cellcolor{gray!15}{}
      &\cellcolor{gray!15}{}
      &\cellcolor{gray!15}{}\\[-9pt]
      \multirow{1}{*}{DenseNet-67}        & \cellcolor{gray!15}{yes} & \cellcolor{gray!15}\textcolor{gray}{\textbf{none}}  & \cellcolor{gray!15}\textcolor{gray}{\textbf{72.3}}\\
      \multirow{1}{*}{%
         \footnotesize{~~-hypercolumn}}   & yes                      & \textbf{Decoder (128 channel)}                      & \textbf{73.6}\\

      \Hline
      \end{tabular}
      \end{small}
      \vspace{-0.01\linewidth}
      \caption{
         \textbf{PASCAL mIoU with COCO pretraining.}
      }
      \label{tab:pascal_coco}
   \end{center}
   \vspace{-0.10\linewidth}
\end{table}

\begin{figure}
   \begin{center}
      \includegraphics[width=0.75\linewidth]{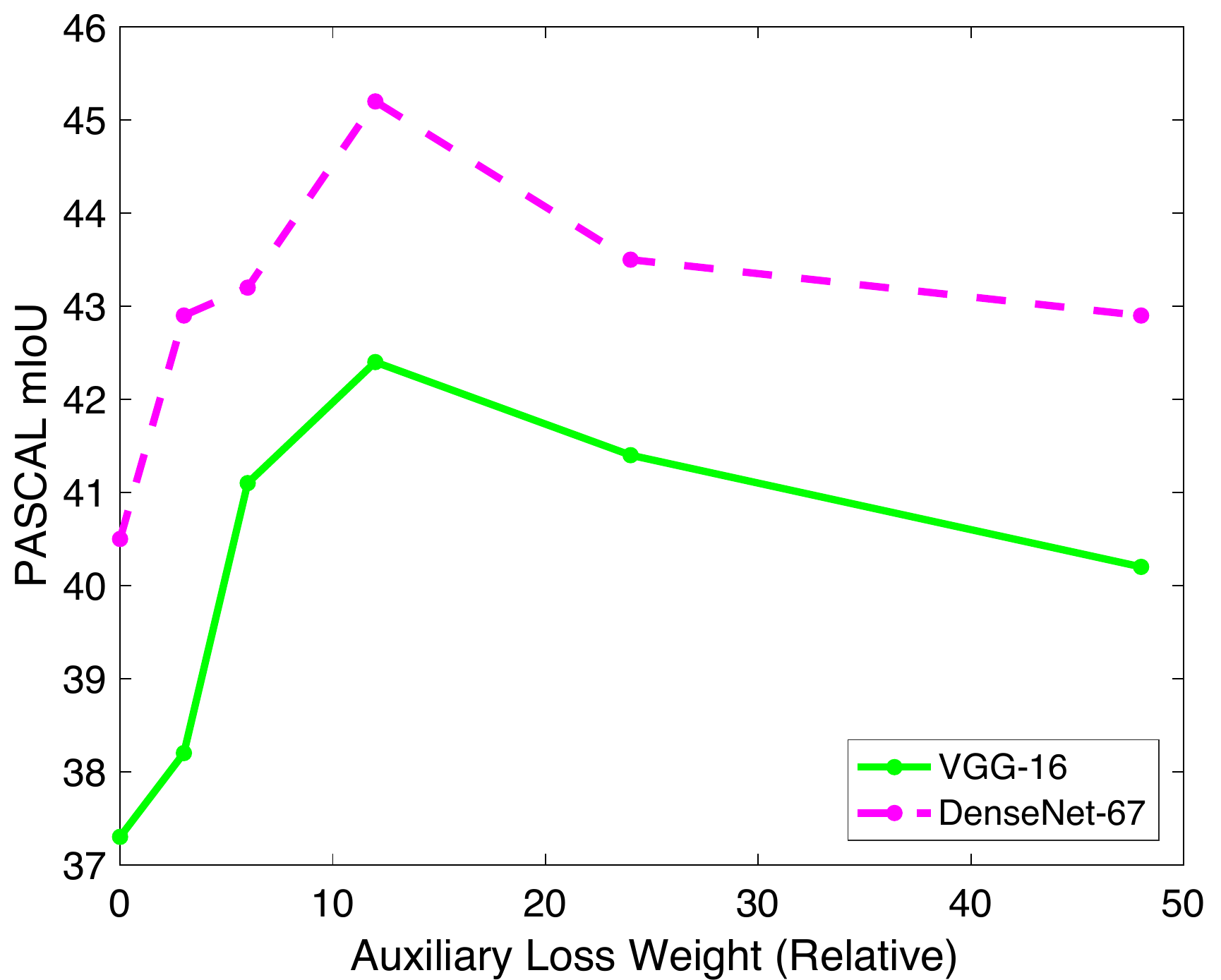}
   \end{center}
   \vspace{-0.03\linewidth}
   \caption{
      \textbf{Auxiliary loss weighting.}
      We plot test performance as a function of the relative weight of the
      losses on the auxiliary vs primary output branches when training with the
      setup in Figure~\ref{fig:overview}.  Weighting is important, but the
      optimal balance appears consistent when changing architecture from VGG-16
      (green) to DenseNet-67 (magenta).  Performance is mIoU on PASCAL, without
      ImageNet pretraining or data augmentation.  Note that any nonzero
      weight on the auxiliary loss (any regularization) improves over
      the baseline.
   }
   \label{fig:weighting}
   \vspace{-0.035\linewidth}
\end{figure}

\begin{figure*}
   \setlength\fboxsep{0pt}
   \begin{center}
   \begin{minipage}[t]{0.98\linewidth}
   \begin{center}
      \begin{minipage}[t]{0.19625\linewidth}
         \vspace{0pt}
         \begin{center}
         \begin{minipage}[t]{0.98\linewidth}
            \vspace{0pt}
            \begin{center}
               \fbox{\includegraphics[width=1.00\linewidth]{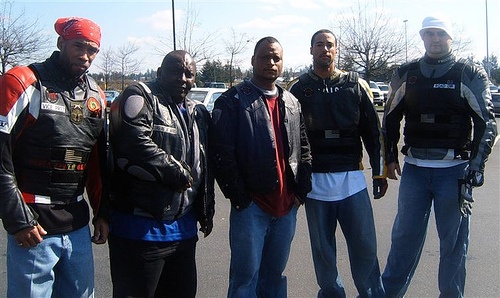}}\\
               \vspace{0.01\linewidth}
               \fbox{\includegraphics[width=1.00\linewidth]{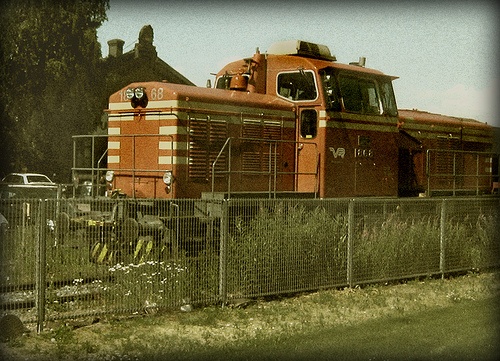}}\\
               \vspace{0.01\linewidth}
               \fbox{\includegraphics[width=1.00\linewidth]{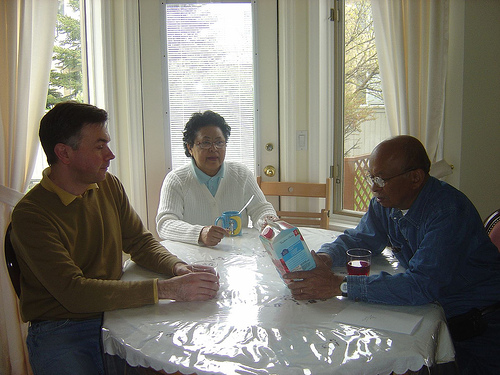}}\\
               \vspace{0.01\linewidth}
               \fbox{\includegraphics[width=1.00\linewidth]{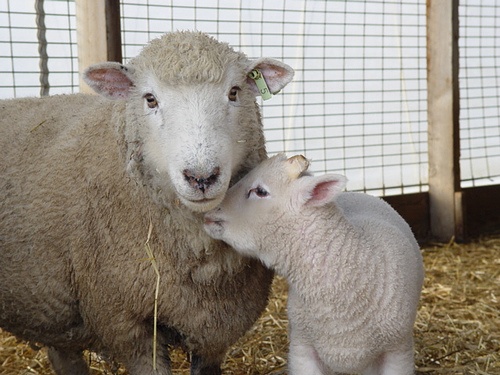}}\\
               \vspace{0.01\linewidth}
               \fbox{\includegraphics[width=1.00\linewidth]{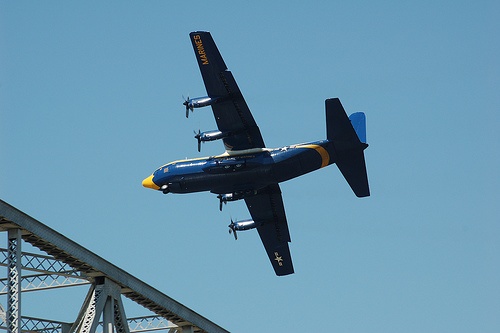}}\\
               \vspace{0.01\linewidth}
               \fbox{\includegraphics[width=1.00\linewidth]{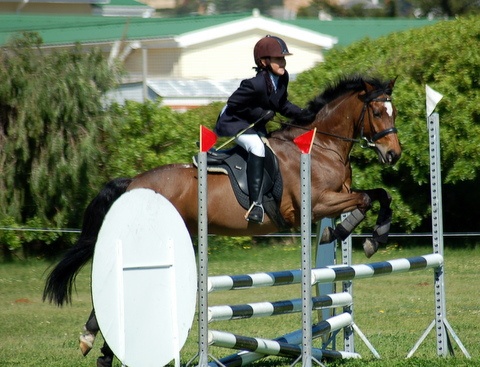}}\\
               \vspace{0.01\linewidth}
               \fbox{\includegraphics[width=1.00\linewidth]{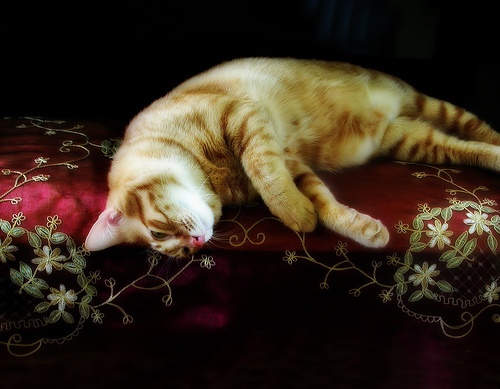}}\\
               \vspace{0.01\linewidth}
               \fbox{\includegraphics[width=1.00\linewidth]{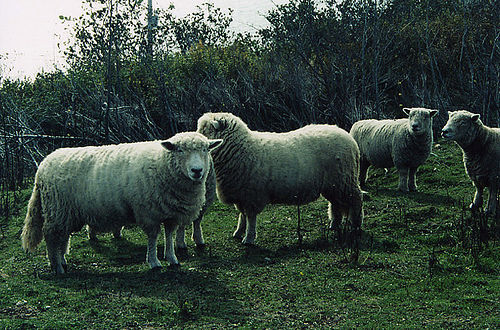}}\\
               \vspace{0.01\linewidth}
               \scriptsize{\textbf{\textsf{Image}}}
            \end{center}
         \end{minipage}
         \end{center}
      \end{minipage}
      \hfill
      \begin{minipage}[t]{0.3925\linewidth}
         \vspace{0pt}
         \begin{center}
         \begin{minipage}[t]{0.49\linewidth}
            \vspace{0pt}
            \begin{center}
               \fbox{\includegraphics[width=1.00\linewidth]{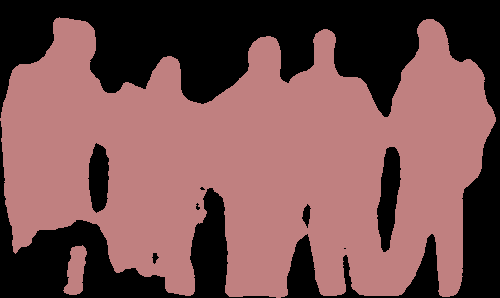}}\\
               \vspace{0.01\linewidth}
               \fbox{\includegraphics[width=1.00\linewidth]{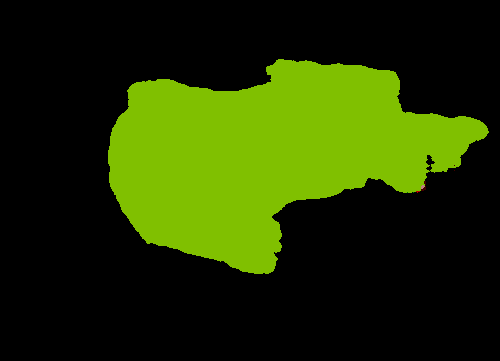}}\\
               \vspace{0.01\linewidth}
               \fbox{\includegraphics[width=1.00\linewidth]{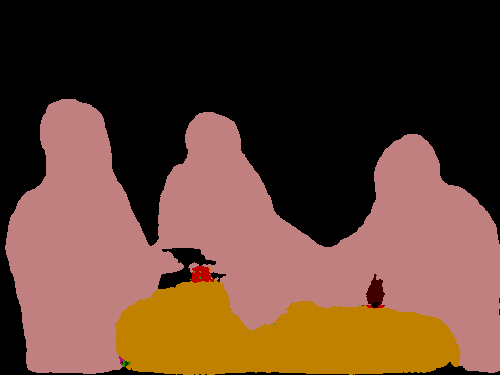}}\\
               \vspace{0.01\linewidth}
               \fbox{\includegraphics[width=1.00\linewidth]{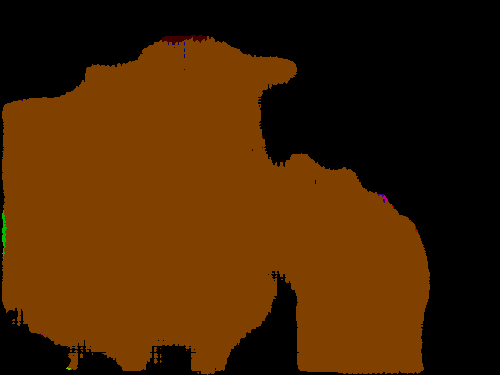}}\\
               \vspace{0.01\linewidth}
               \fbox{\includegraphics[width=1.00\linewidth]{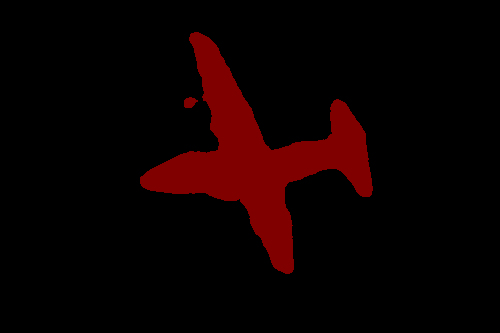}}\\
               \vspace{0.01\linewidth}
               \fbox{\includegraphics[width=1.00\linewidth]{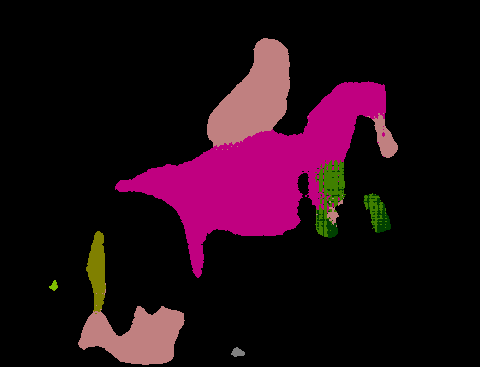}}\\
               \vspace{0.01\linewidth}
               \fbox{\includegraphics[width=1.00\linewidth]{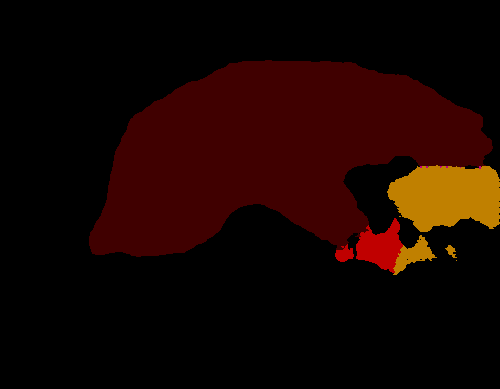}}\\
               \vspace{0.01\linewidth}
               \fbox{\includegraphics[width=1.00\linewidth]{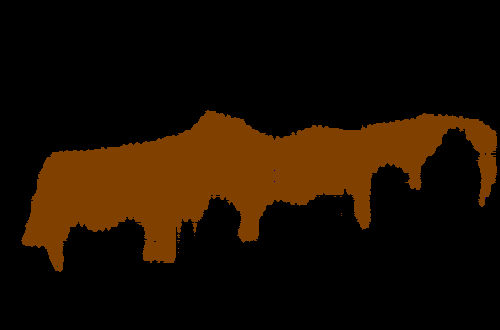}}\\
               \vspace{0.01\linewidth}
               \scriptsize{\textbf{\textsf{Auxiliary Output}}}
            \end{center}
         \end{minipage}
         \begin{minipage}[t]{0.49\linewidth}
            \vspace{0pt}
            \begin{center}
               \fcolorbox{red}{white}{\includegraphics[width=1.00\linewidth]{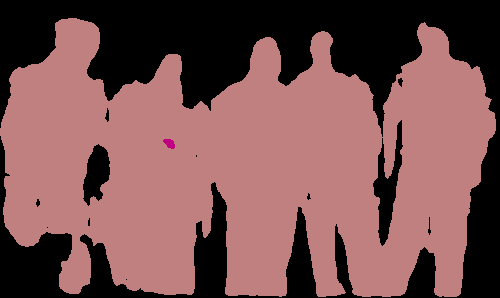}}\\
               \vspace{0.01\linewidth}
               \fcolorbox{red}{white}{\includegraphics[width=1.00\linewidth]{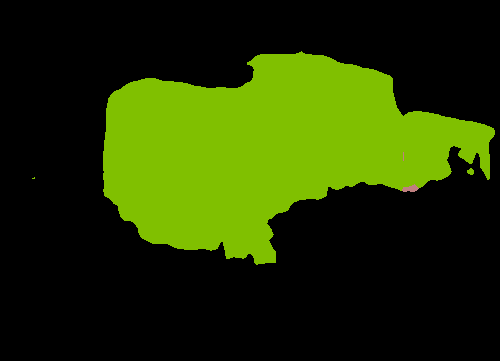}}\\
               \vspace{0.01\linewidth}
               \fcolorbox{red}{white}{\includegraphics[width=1.00\linewidth]{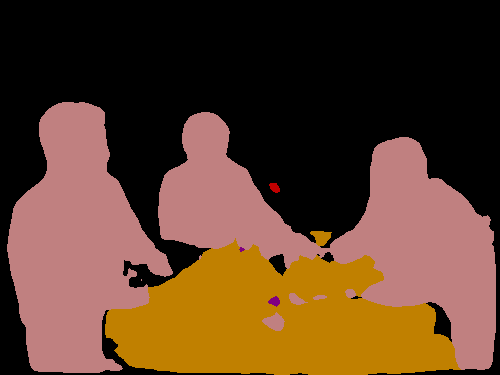}}\\
               \vspace{0.01\linewidth}
               \fcolorbox{red}{white}{\includegraphics[width=1.00\linewidth]{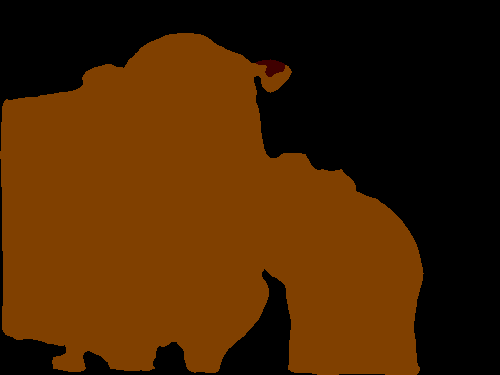}}\\
               \vspace{0.01\linewidth}
               \fcolorbox{red}{white}{\includegraphics[width=1.00\linewidth]{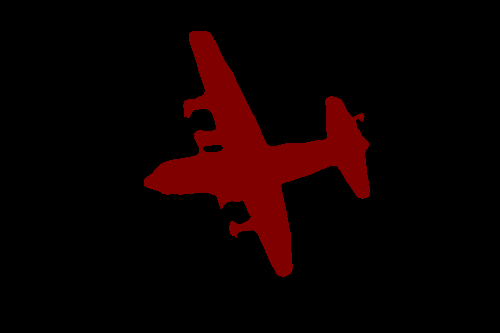}}\\
               \vspace{0.01\linewidth}
               \fcolorbox{red}{white}{\includegraphics[width=1.00\linewidth]{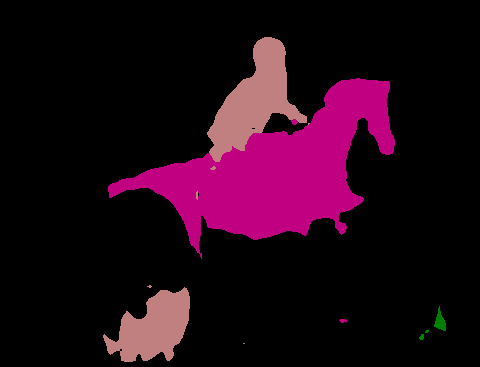}}\\
               \vspace{0.01\linewidth}
               \fcolorbox{red}{white}{\includegraphics[width=1.00\linewidth]{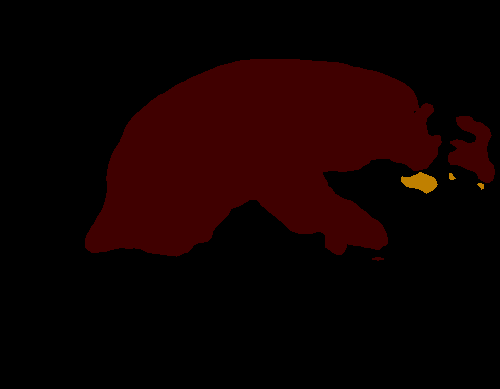}}\\
               \vspace{0.01\linewidth}
               \fcolorbox{red}{white}{\includegraphics[width=1.00\linewidth]{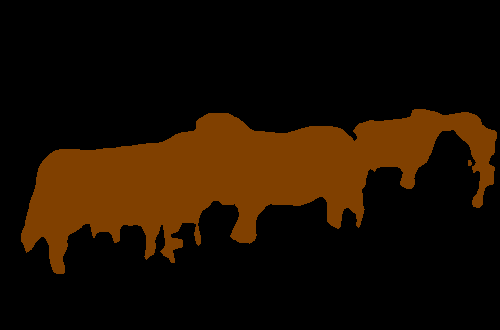}}\\
               \vspace{0.01\linewidth}
               \scriptsize{\textbf{\textsf{\textcolor{red}{Primary Output}}}}
            \end{center}
         \end{minipage}\\
         \vspace{1pt}
         \rule{0.975\linewidth}{0.5pt}\\
         \scriptsize{\textbf{\textsf{Our System: DenseNet-67 trained with regularizer}}}
         \end{center}
      \end{minipage}
      \hfill
      \begin{minipage}[t]{0.19625\linewidth}
         \vspace{0pt}
         \begin{center}
         \begin{minipage}[t]{0.98\linewidth}
            \vspace{0pt}
            \begin{center}
               \fbox{\includegraphics[width=1.00\linewidth]{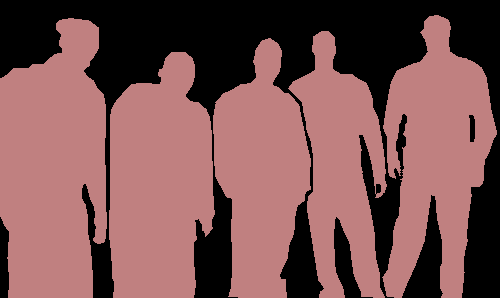}}\\
               \vspace{0.01\linewidth}
               \fbox{\includegraphics[width=1.00\linewidth]{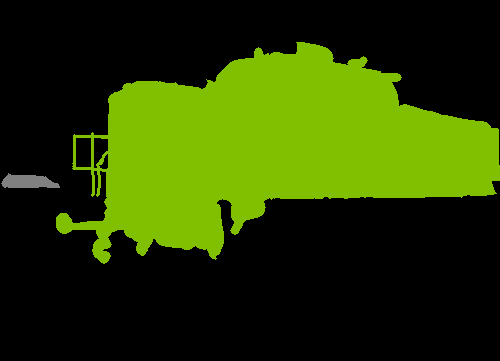}}\\
               \vspace{0.01\linewidth}
               \fbox{\includegraphics[width=1.00\linewidth]{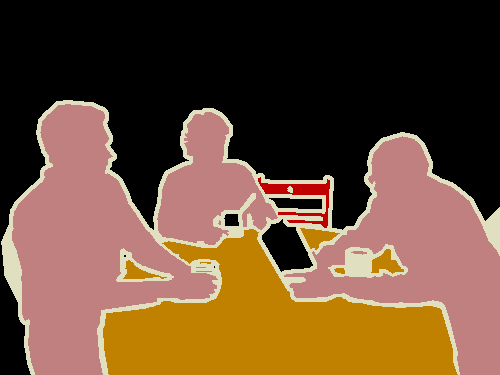}}\\
               \vspace{0.01\linewidth}
               \fbox{\includegraphics[width=1.00\linewidth]{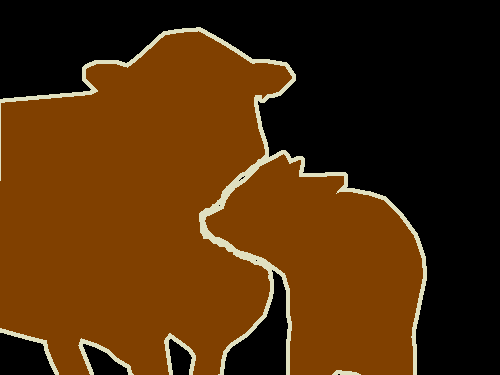}}\\
               \vspace{0.01\linewidth}
               \fbox{\includegraphics[width=1.00\linewidth]{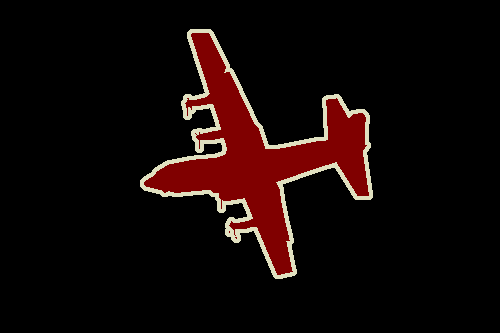}}\\
               \vspace{0.01\linewidth}
               \fbox{\includegraphics[width=1.00\linewidth]{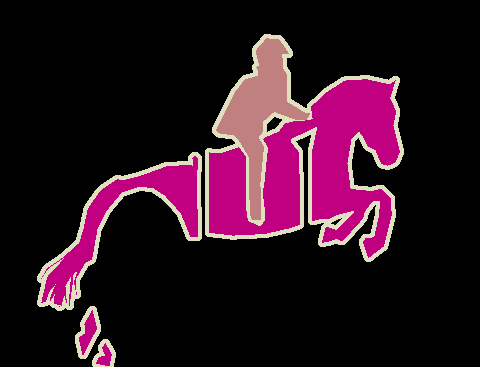}}\\
               \vspace{0.01\linewidth}
               \fbox{\includegraphics[width=1.00\linewidth]{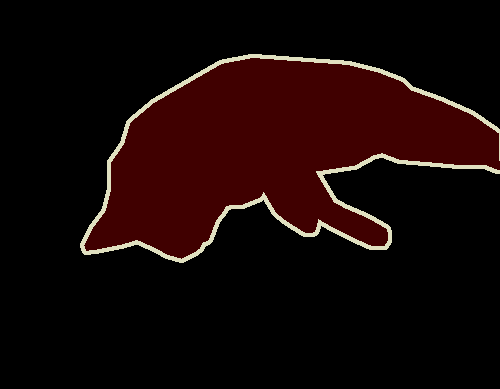}}\\
               \vspace{0.01\linewidth}
               \fbox{\includegraphics[width=1.00\linewidth]{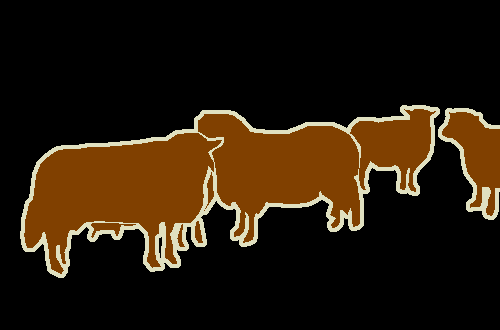}}\\
               \vspace{0.01\linewidth}
               \scriptsize{\textbf{\textsf{Ground-truth}}}
            \end{center}
         \end{minipage}
         \end{center}
      \end{minipage}
      \hfill
      \begin{minipage}[t]{0.19625\linewidth}
         \vspace{0pt}
         \begin{center}
         \begin{minipage}[t]{0.98\linewidth}
            \vspace{0pt}
            \begin{center}
               \fbox{\includegraphics[width=1.00\linewidth]{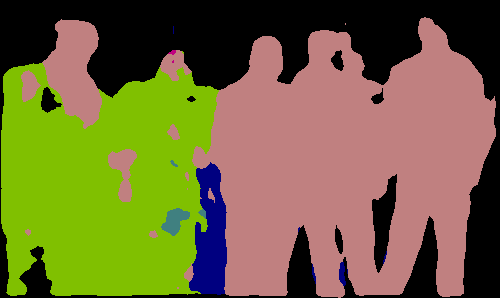}}\\
               \vspace{0.01\linewidth}
               \fbox{\includegraphics[width=1.00\linewidth]{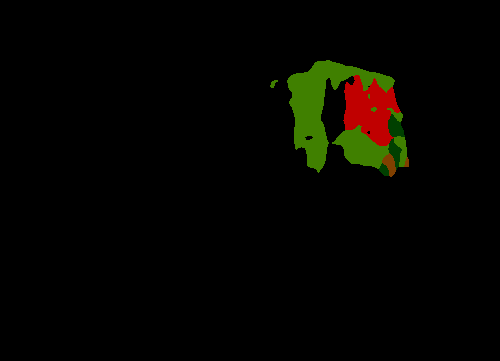}}\\
               \vspace{0.01\linewidth}
               \fbox{\includegraphics[width=1.00\linewidth]{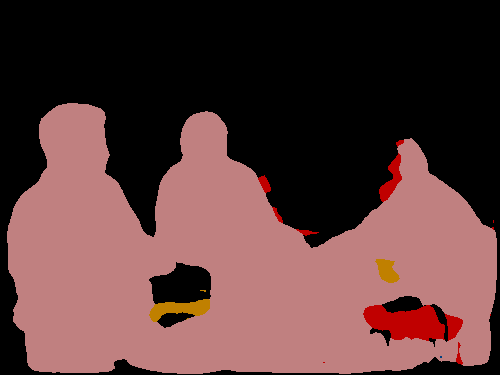}}\\
               \vspace{0.01\linewidth}
               \fbox{\includegraphics[width=1.00\linewidth]{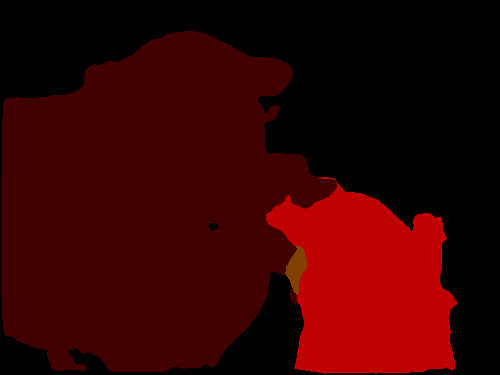}}\\
               \vspace{0.01\linewidth}
               \fbox{\includegraphics[width=1.00\linewidth]{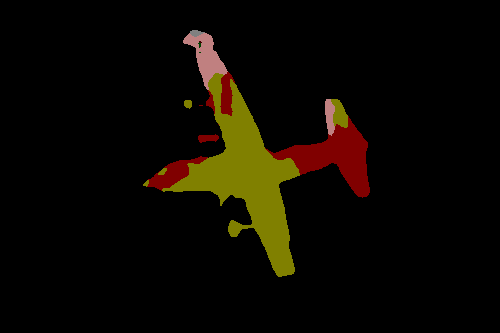}}\\
               \vspace{0.01\linewidth}
               \fbox{\includegraphics[width=1.00\linewidth]{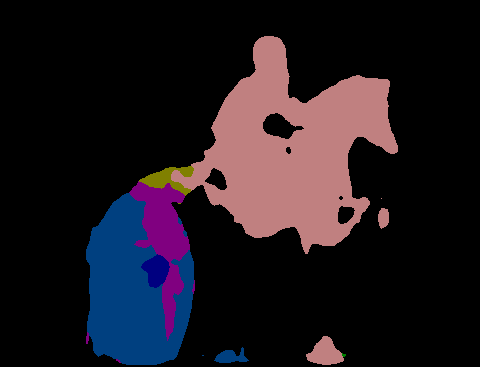}}\\
               \vspace{0.01\linewidth}
               \fbox{\includegraphics[width=1.00\linewidth]{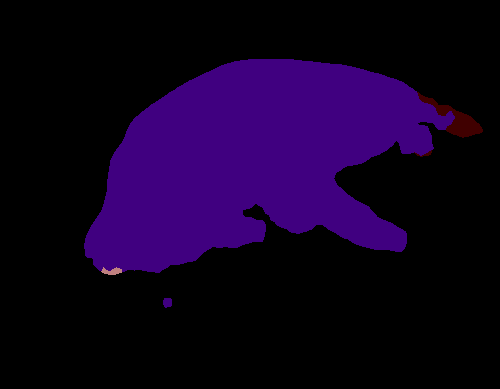}}\\
               \vspace{0.01\linewidth}
               \fbox{\includegraphics[width=1.00\linewidth]{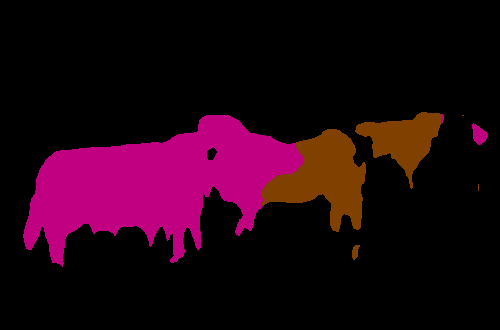}}\\
               \vspace{0.01\linewidth}
               \scriptsize{\textbf{\textsf{Baseline DenseNet-67}}}
            \end{center}
         \end{minipage}
         \end{center}
      \end{minipage}
   \end{center}
   \end{minipage}
   \end{center}
   \vspace{-0.01\linewidth}
   \caption{
      \textbf{Semantic segmentation results on PASCAL.}
      We show the output of a baseline 67-layer hypercolumn DenseNet (rightmost
      column) compared to that of the same architecture trained with our
      auxiliary decoder branch as a regularizer (middle columns).  All examples
      are from the validation set.  While we can discard the auxiliary branch
      after training, we include its output here to display the decoder's
      operation.  Our network provides high-level signals to the decoder which,
      in turn, produces reasonable segmentations.  To best illustrate the
      effect of regularization, all results shown are for networks trained from
      scratch, without ImageNet pretraining or data augmentation.  This
      corresponds to the $40.5$ to $45.2$ jump in mIoU reported in
      Table~\ref{tab:pascal_scratch}, between the baseline and our primary
      output.
   }
   \label{fig:pascal}
\end{figure*}

\subsection{Semantic Segmentation Results}
\label{sec:results}

Tables~\ref{tab:pascal_scratch},~\ref{tab:pascal_imagenet},
and~\ref{tab:pascal_coco} summarize the performance benefits of training with
our regularizer.  In the absence of pretraining or data augmentation, we boost
performance of both VGG-16 and DenseNet-67 by $5.1$ and $4.7$ mIoU,
respectively, which is more than a $10\%$ relative boost.  Regularization with
our decoder still improves mIoU (from $58.8$ to $60.6$) of DenseNet-67 trained
with data augmentation.  To further show the robustness of our regularization
scheme to the choice of architecture, we also experiment with an FCN~\cite{FCN}
version of VGG-16, as included in Table~\ref{tab:pascal_scratch}.

Table~\ref{tab:pascal_ablation} demonstrates the necessity of our two-phase
training procedure.  If we unfreeze the decoder and update its parameters in
the second training phase, test performance of the primary output deteriorates.
Likewise, if we skip the first phase, and train from scratch with an unfrozen,
randomly initialized decoder, the accuracy gain disappears.  Thus, the
regularization effect is due to a transfer of information from the learned
label model, rather than stemming from an architectural design of dual output
pathways.

Table~\ref{tab:pascal_imagenet} shows that our regularization scheme
synergizes with ImageNet pretraining.  It improves VGG-16 performance, and
even provides some benefit to a very deep 121-layer DenseNet pretrained on
ImageNet, while using data augmentation.  A baseline $71.6$ mIoU for DenseNet
appears near state-of-the-art for networks that do not employ additional tricks
(\eg custom pooling layers~\cite{PSPNet}, use of multiscale, or post-processing
with CRFs~\cite{DeepLab}).  Our improvement to $71.9$ mIoU may be nontrivial.
Expanding trials in combination with pretraining, our regularizer improves
results when pretraining on COCO, as shown in Table~\ref{tab:pascal_coco}.

We also combine our regularizer with the latest network design for semantic
segmentation: dilated ResNet augmented with the pyramid pooling module of
PSPNet~\cite{PSPNet}.  We used the output of the pyramid pooling layer to
predict input for the decoder and semantic segmentation.
Table~\ref{tab:pascal_imagenet} shows gain over the corresponding PSPNet
baseline.

Beyond autoencoder architecture choice, application of our regularizer involves
one free parameter: the relative weight of the auxiliary branch loss
with respect to the primary loss.  Figure~\ref{fig:weighting} shows how
performance of the trained network varies with this parameter, when using our
32-channel bottleneck layer decoder with MSE loss on the auxiliary branch.

We have also run similar experiments with cross-entropy loss on the auxiliary
branch with the weight parameter in $[0,6]$.  Here, the weight parameter range
is changed due to the difference in the dynamic range of values between MSE
loss and cross-entropy loss.  Behaving similarly to Figure~\ref{fig:weighting},
relative weighting of $0.5$ achieves the highest accuracy.  We use this weight
value across all of the experiments using our decoder with 128-channel
bottleneck layer.  While the regularizer always provides a benefit, placing a
proper relative weight on the auxiliary loss is important.

Figure~\ref{fig:pascal} visualizes the impact of training with our learned
label decoder as a regularizer.  Most notably, the network trained with
regularization appears to correct some global or large-scale semantic errors
in comparison to the baseline.  Contrast such behavior to CRF-based
post-processing, which typically achieves impact through fixing local mistakes.
Also notable is that our auxiliary output itself is quite reasonable.  This
suggests that the autoencoder training phase is successful in creating encoders
and decoders that model label structure.

\subsection{Label Model Introspection}
\label{sec:introspection}

\begin{figure*}
   \setlength{\fboxsep}{0pt}
   \setlength{\fboxrule}{1pt}
   \begin{center}
   \begin{minipage}[t]{0.49\linewidth}
      \vspace{0pt}
      \fcolorbox{red!10}{red!10}{
      \begin{tabular}{ccc}
      \vspace{-5pt}
      {\fcolorbox{red!10}{red!10}{~}}\\
      \hspace{-4.5pt}{\fcolorbox{mediumgreen}{white}{\includegraphics[width=2.55cm]{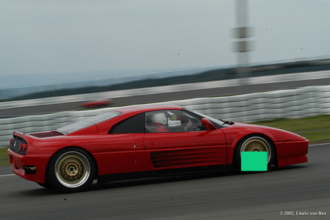}}}&\hspace{-10pt}
      {\fbox{\includegraphics[width=2.55cm]{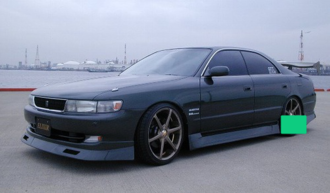}}}&\hspace{-10pt}
      {\fbox{\includegraphics[width=2.55cm]{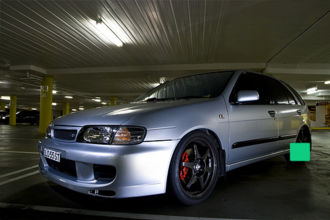}}}\\
      \vspace{3pt}
      \hspace{-4.5pt}{\fcolorbox{mediumgreen}{white}{\includegraphics[width=2.55cm]{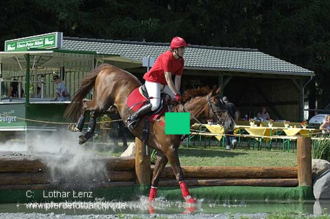}}}&\hspace{-10pt}
      {\fbox{\includegraphics[width=2.55cm]{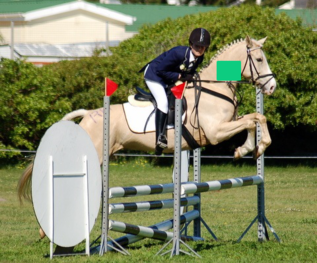}}}&\hspace{-10pt}
      {\fbox{\includegraphics[width=2.55cm]{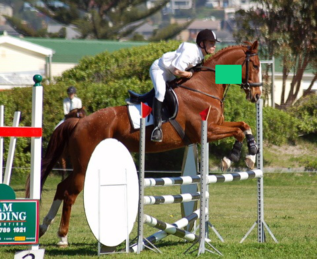}}}\\
      \vspace{3pt}
      \hspace{-4.5pt}{\fcolorbox{mediumgreen}{white}{\includegraphics[width=2.55cm]{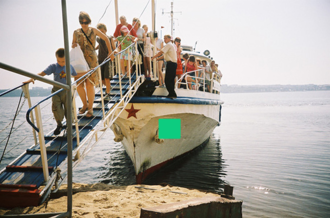}}}&\hspace{-10pt}
      {\fbox{\includegraphics[width=2.55cm]{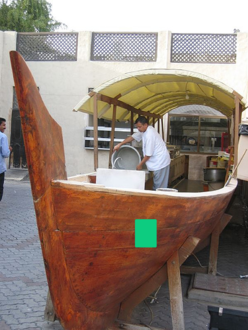}}}&\hspace{-10pt}
      {\fbox{\includegraphics[width=2.55cm]{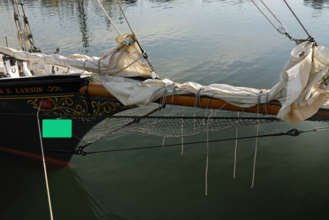}}}\\
      \vspace{3pt}
      \hspace{-4.5pt}{\fcolorbox{mediumgreen}{white}{\includegraphics[width=2.55cm]{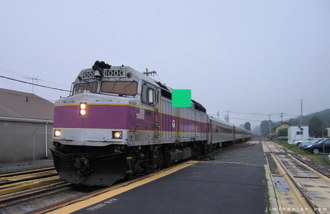}}}&\hspace{-10pt}
      {\fbox{\includegraphics[width=2.55cm]{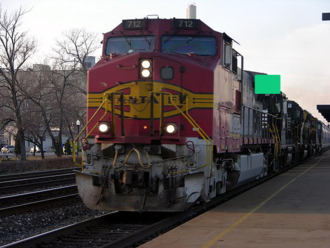}}}&\hspace{-10pt}
      {\fbox{\includegraphics[width=2.55cm]{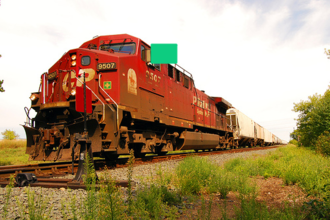}}}\\
      \vspace{3pt}
      \hspace{-4.5pt}{\fcolorbox{mediumgreen}{white}{\includegraphics[width=2.55cm]{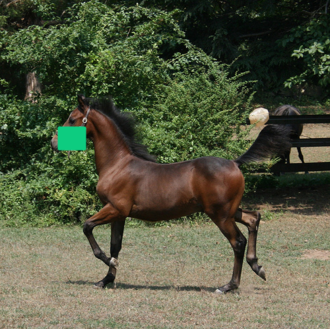}}}&\hspace{-10pt}
      {\fbox{\includegraphics[width=2.55cm]{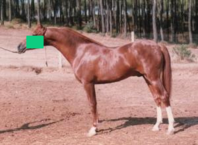}}}&\hspace{-10pt}
      {\fbox{\includegraphics[width=2.55cm]{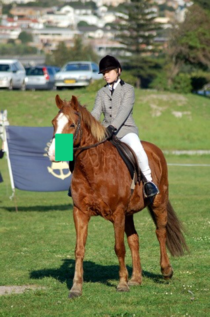}}}\\
      \vspace{3pt}
      \hspace{-4.5pt}{\fcolorbox{mediumgreen}{white}{\includegraphics[width=2.55cm]{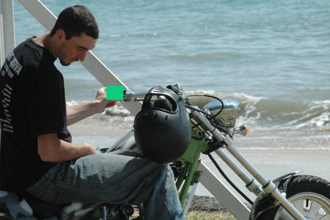}}}&\hspace{-10pt}
      {\fbox{\includegraphics[width=2.55cm]{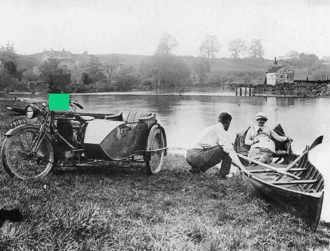}}}&\hspace{-10pt}
      {\fbox{\includegraphics[width=2.55cm]{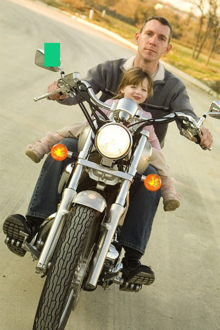}}}\\
      \vspace{3pt}
      \hspace{-4.5pt}{\fcolorbox{mediumgreen}{white}{\includegraphics[width=2.55cm]{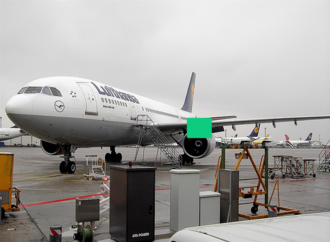}}}&\hspace{-10pt}
      {\fbox{\includegraphics[width=2.55cm]{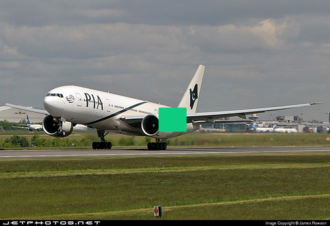}}}&\hspace{-10pt}
      {\fbox{\includegraphics[width=2.55cm]{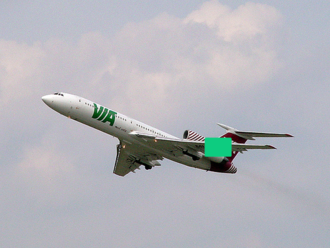}}}
      \end{tabular}}
   \end{minipage}
   \hfill
   \begin{minipage}[t]{0.49\linewidth}
      \vspace{15pt}
      \fcolorbox{gray!20}{gray!20}{%
      \begin{tabular}{ccc}
      \vspace{-5pt}
      {\fcolorbox{gray!20}{gray!20}{~}}\\
      \hspace{-4.5pt}{\fcolorbox{mediumgreen}{white}{\includegraphics[width=2.55cm]{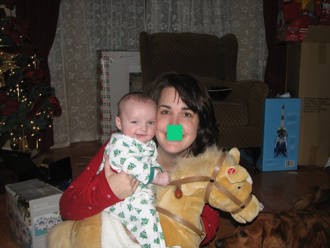}}}&\hspace{-10pt}
      {\fbox{\includegraphics[width=2.55cm]{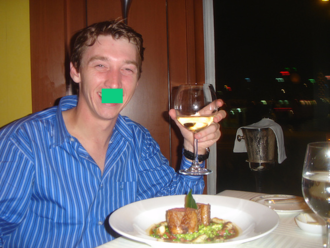}}}&\hspace{-10pt}
      {\fbox{\includegraphics[width=2.55cm]{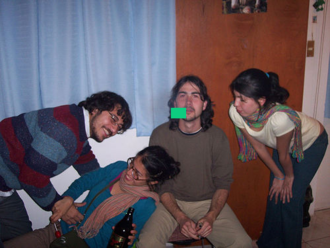}}}\\
      \vspace{3pt}
      \hspace{-4.5pt}{\fcolorbox{mediumgreen}{white}{\includegraphics[width=2.55cm]{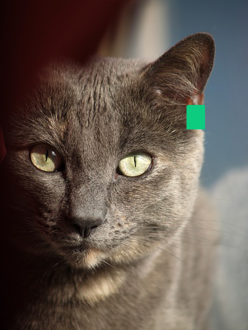}}}&\hspace{-10pt}
      {\fbox{\includegraphics[width=2.55cm]{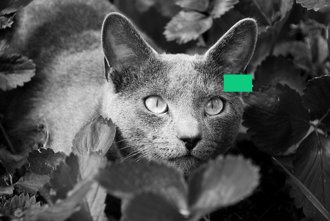}}}&\hspace{-10pt}
      {\fbox{\includegraphics[width=2.55cm]{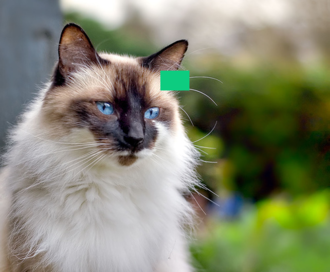}}}\\
      \vspace{3pt}
      \hspace{-4.5pt}{\fcolorbox{mediumgreen}{white}{\includegraphics[width=2.55cm]{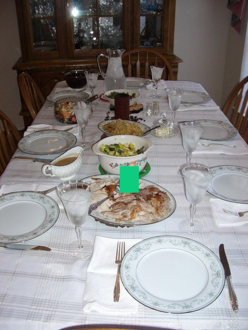}}}&\hspace{-10pt}
      {\fbox{\includegraphics[width=2.55cm]{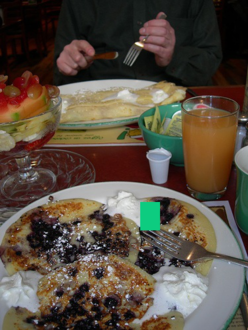}}}&\hspace{-10pt}
      {\fbox{\includegraphics[width=2.55cm]{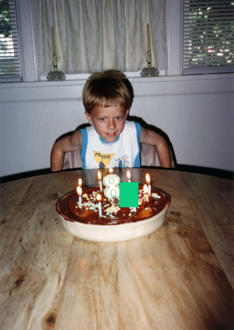}}}\\
      \vspace{3pt}
      \hspace{-4.5pt}{\fcolorbox{mediumgreen}{white}{\includegraphics[width=2.55cm]{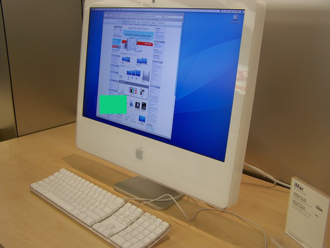}}}&\hspace{-10pt}
      {\fbox{\includegraphics[width=2.55cm]{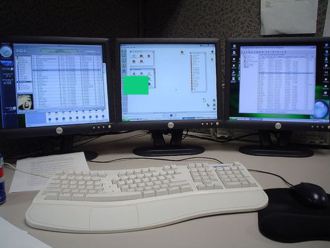}}}&\hspace{-10pt}
      {\fbox{\includegraphics[width=2.55cm]{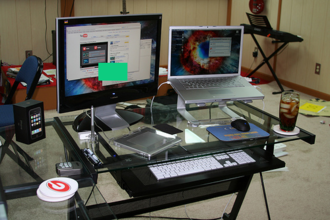}}}
      \end{tabular}}
      ~\\~\\~\\
      \fcolorbox{red!10}{red!10}{%
      \begin{tabular}{ccc}
      \vspace{-5pt}
      {\fcolorbox{red!10}{red!10}{~}}\\
      \hspace{-4.5pt}{\fcolorbox{mediumgreen}{white}{\includegraphics[width=2.55cm]{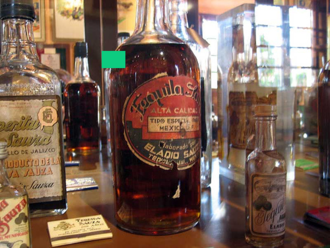}}}&\hspace{-10pt}
      {\fbox{\includegraphics[width=2.55cm]{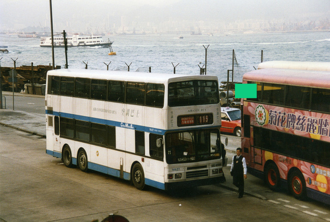}}}&\hspace{-10pt}
      {\fbox{\includegraphics[width=2.55cm]{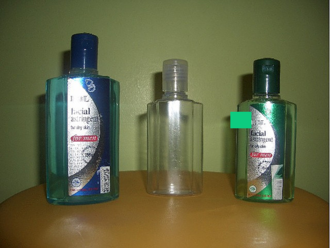}}}\\
      \vspace{3pt}
      \hspace{-4.5pt}{\fcolorbox{mediumgreen}{white}{\includegraphics[width=2.55cm]{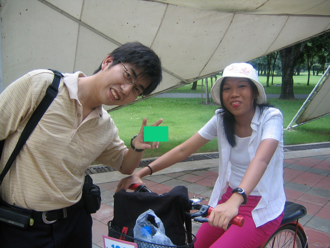}}}&\hspace{-10pt}
      {\fbox{\includegraphics[width=2.55cm]{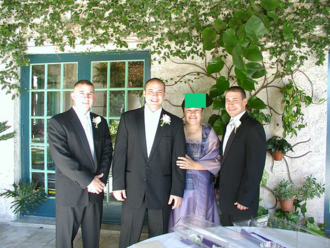}}}&\hspace{-10pt}
      {\fbox{\includegraphics[width=2.55cm]{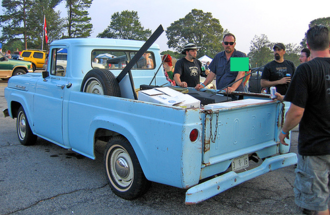}}}\\
      \vspace{3pt}
      \hspace{-4.5pt}{\fcolorbox{mediumgreen}{white}{\includegraphics[width=2.55cm]{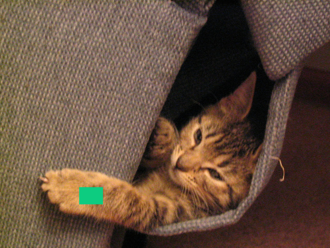}}}&\hspace{-10pt}
      {\fbox{\includegraphics[width=2.55cm]{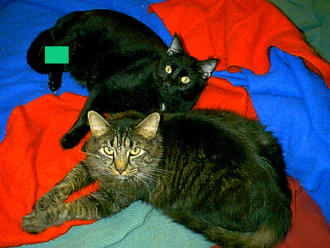}}}&\hspace{-10pt}
      {\fbox{\includegraphics[width=2.55cm]{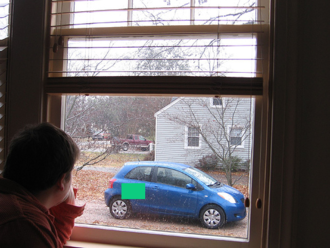}}}
      \end{tabular}}
   \end{minipage}
   \end{center}
   \vspace{-0.01\linewidth}
   \caption{
      \textbf{Finding regions with similar representations.}
      For each \textcolor{mediumgreen}{\textbf{query}} image (green border) and
      region (green dot), the next two images to the right are those in the
      validation set containing the nearest regions to the query region.  All
      query images are from the training set.  For examples on red background,
      search is conducted not by looking at images, but via matching features
      produced by the \textcolor{red}{\textbf{encoder}} run on ground-truth
      label maps.  The bottom-right shows failure cases, such as matching a
      cat's arm to the car rear door.  For examples on gray background, our
      \textcolor{darkgray}{\textbf{DenseNet-67-hypercolumn CNN}} is used to
      predict the label space search representations from images.
   }
   \label{fig:nearest}
\end{figure*}

To further investigate what the autoencoder learns, we consider using the
bottleneck representation produced by the encoder as defining features by
which we can perform queries in label space.  Specifically, we pick a region
of a training image label and represent that region with features extracted
from bottleneck layer.  As the bottleneck layer is low resolution, we are
selecting features at coarse, but corresponding spatial location.

Next, we perform nearest neighbor search over all regions in the validation
set and find the two closest regions to the query region.
Figure~\ref{fig:nearest} shows the results of this experiment.  Returned
regions not only have the same object class types as the query regions, but
also share similar shapes to that of the query.  This reveals that our
label autoencoder has learned to capture object shape characteristics.

We also repeat this experiment, except with queries starting from images.
Here the bottleneck representation is produced by a CNN, which was trained with
both hypercolumn and decoder prediction pathways; the latter yields the
required features.  As shown in the top-right of Figure~\ref{fig:nearest},
returned regions have similar context and shape to the query.

\section{Conclusion}
\label{sec:conclusion}

Our novel regularization method, when applied to training deep networks for
semantic segmentation, consistently improves their generalization performance.
The intuition behind our work, that additional supervisory signal can be
squeezed from highly detailed annotation, is supported by the types of errors
this regularizer corrects, as well as our efforts at introspection into our
learned label model.

Our results also indicate that one should now reevaluate the relative utility
of different forms of annotation; our method makes detailed labeling more
useful than previously believed.  This observation may be especially important
for applications of computer vision, such as self-driving cars, that demand
detailed scene understanding, and for which large-scale dataset construction is
essential.

~\\
\noindent
{\small
\textbf{Acknowledgements.}  This work was in part supported by the DARPA
Lifelong Learning Machines program.}

{\small
\bibliographystyle{ieee}
\bibliography{label-reg_arxiv}
}

\end{document}